\documentclass[10pt,twocolumn,letterpaper]{article}

\usepackage[pagenumbers]{cvpr} 

\usepackage[dvipsnames]{xcolor}

\definecolor{cvprblue}{rgb}{0.21,0.49,0.74}
\usepackage[pagebackref,breaklinks,colorlinks,citecolor=cvprblue]{hyperref}
\usepackage{xcolor}
\usepackage{multirow}
\usepackage[switch]{lineno}
\usepackage{colortbl}

\def\paperID{133} 
\def\confName{3DV\xspace}
\def\confYear{2025\xspace}

\title{JADE: Joint-aware Latent Diffusion for 3D Human Generative Modeling}

\author{Haorui Ji \quad Rong Wang \quad Tao Jun Lin \quad Hongdong Li\\
The Australian National University\\
{\tt\small \{haorui.ji, rong.wang, taojun.lin, hongdong.li\}@anu.edu.au}
}

\begin{document}
\maketitle
\begin{abstract}

Generative modeling of 3D human bodies have been studied extensively in computer vision. The core is to design a compact latent representation that is both expressive and semantically interpretable, yet existing approaches struggle to achieve both requirements. In this work, we introduce JADE, a generative framework that learns the variations of human shapes with fined-grained control. Our key insight is a joint-aware latent representation that decomposes human bodies into skeleton structures, modeled by joint positions, and local surface geometries, characterized by features attached to each joint. This disentangled latent space design enables geometric and semantic interpretation, facilitating users with flexible controllability. To generate coherent and plausible human shapes under our proposed decomposition, we also present a cascaded pipeline where two diffusions are employed to model the distribution of skeleton structures and local surface geometries respectively. Extensive experiments are conducted on public datasets, where we demonstrate the effectiveness of JADE framework in multiple tasks in terms of autoencoding reconstruction accuracy, editing controllability and generation quality compared with existing methods.

\end{abstract}    
\section{Introduction}
\label{sec:intro}

Generative modeling of 3D human bodies is key to many practical applications, such as multimedia, healthcare, virtual and augmented reality \cite{bridgeman2019multi,jiang2022golfpose,stenum2021applications}. To better facilitates these applications, the human model should achieve two main goals: (i) it can accurately encode body shape details to generate high-quality samples without artifacts (ii) it enables join-level fine-grained manipulation, while producing realistic and plausible pose-dependent deformations. However, existing works mostly fail to meet both requirements.  \cite{loper2023smpl,pavlakos2019expressive,anguelov2005scape} adopt a statistical model with a linear parametric space to encode body shapes, which is not sufficiently expressive thus resulting in degraded. While recent learning-based methods~\cite{tretschk2020demea,jiang2020disentangled,xu2020ghum} improves reconstruction quality, their latent encodings are not disentangled, making them hard to flexibly edit the human bodies.

\begin{figure}[t]
    \begin{center}
    \includegraphics[width=\linewidth]{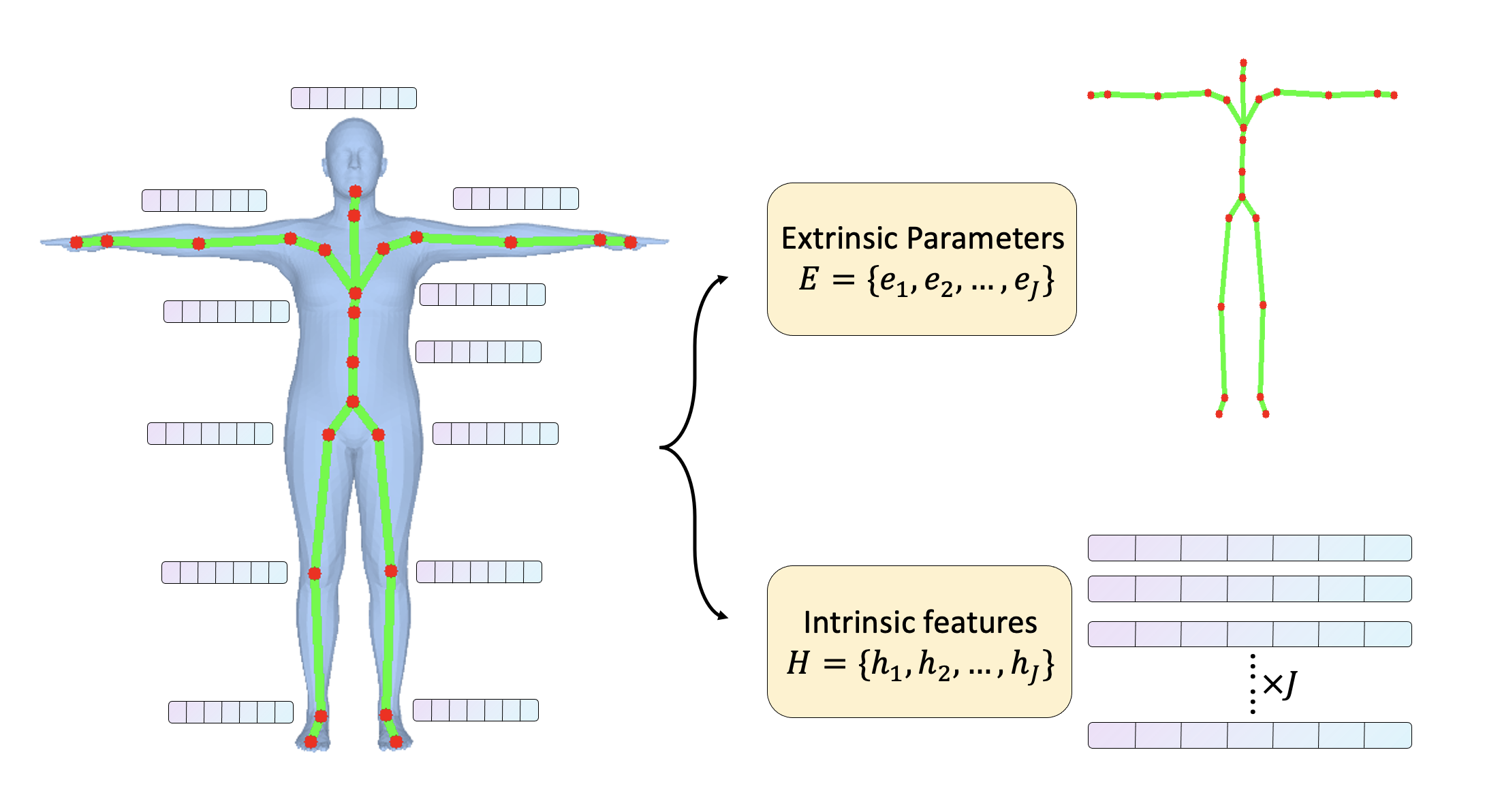}
    \caption{Overview of our joint-aware latent representation. We dispatch modeling of the overall human body into a sequence of \emph{joint tokens}, where each token contains extrinsic parameters for encoding skeleton structures, and intrinsic features for modeling local surface geometries.}
    \label{fig:joint-aware decomposition}
    \end{center}
\vspace{-1.5em}
\end{figure}

In this work, we introduce \textbf{J}oint-aware L\textbf{A}tent \textbf{D}iffusion for 3D Human G\textbf{E}nerative Modeling (\textbf{JADE}), a new method that satisfy both the expressiveness and controllability requirements in a principled way by building a diffusion-based generative model that operates on human body surface point cloud. Firstly, unlike previous works that often represent human bodies as a holistic latent code~\cite{loper2023smpl,pavlakos2019expressive}, we introduce a joint-aware latent representation by dispatching body encoding into individual joints, so that the overall human model can be described as a sequence of \emph{joint tokens}, akin to the tokenization process in natural language processing~\cite{vaswani2017attention}. Each joint token are further split into extrinsic parameters and intrinsic features, where the former encodes body kinematic configurations and pose-dependent deformations, and the latter encodes canonical body shapes through local surface geometries, as shown in Figure \ref{fig:joint-aware decomposition}. This disentangled joint-level representation naturally allows local sampling and editing human parts associated to each joint, thus are more flexible and controllable. Furthermore, to capture joint-wise correlation, we utilize a Transformer-based architecture~\cite{dosovitskiy2020image} to fuse features across joints via attention, which enhances model expressiveness and ensures consistency between articulated joints.

We also develop a cascaded diffusion pipeline that allows for better modeling the latent distribution under our proposed factorization so that we can generate plausible human shapes through independently sampled tokens. Leveraging the fact that joint-wise tokens can be factorized into extrinsic-intrinsic pairs, our pipeline learns two diffusions successively, one generating extrinsic parameters to explicitly express any spatial and structural information, while the other producing the intrinsic features conditioned on the extrinsics to supplement local geometric details and improves the generation quality. This design allows each generated surface point to be informed of both global skeletal information as well as individual local structure style, enriching the synthesized geometry with useful structural information and allowing for fine-grained manipulations.

To summarize, our main contributions are as follows:
\begin{itemize}
\item We propose a joint-aware latent representation for generative 3D human modeling that encode both the skeleton information as well as local structure geometric details.

\item We propose a two-phase cascaded diffusion pipeline to effectively model the distribution of disentangled latent space and generate plausible human shapes through independently sampled tokens.

\item We showcase that our method not only allows for generating high-fidelity and diverse human shapes, but also enables multiple shape editing and variation applications, demonstrating its effectiveness.
\end{itemize}

\section{Related Work}
\label{sec:related work}
\subsection{Human Body Modeling}
Human body modeling refers to learning a statistical model by exploiting extensive collection of 3D body scans with or without hand articulation and facial expression. Classical parametric models like SMPL, SCAPE, etc.~\cite{loper2023smpl,pavlakos2019expressive,anguelov2005scape}, factor full body deformations into identity-dependent and pose-dependent components, accompanied with a skinning algorithm that deforms surface vertices as a function of underlying skeleton change. Despite success in various applications, they still possess inherent drawbacks. For example, their expressivity is limited by the linearity of PCA subspace and struggle to represent highly nonlinear soft-tissue deformations. Moreover, fully separation of shape and pose parameters may not capture the interdependencies between body shape changes and specific poses, leading to less accurate representations in dynamic scenarios.

Recent works~\cite{jiang2020disentangled,bouritsas2019neural,tretschk2020demea,xu2020ghum,zhou2020unsupervised,osman2022supr,sun2023learning} have started to adopt deep learning-based approaches on polygonal meshes of human bodies. By constructing different convolution-like operators for feature extraction on irregular meshes, they model articulations as vertex offsets field warped from template mesh in canonical space to deformed space. However, most of them still use single latent code to represent global shape variations. Not only does it tend to capture spurious long-range correlations and produce non-local deformation artifacts, but also makes it hard to conduct intuitive editing since the semantics of the latent is vague. Among the few who construct structure-level latents~\cite{jiang2020disentangled,sun2023learning}, a common practice is to partition human bodies into anatomical parts, assign each surface vertex a hard segmentation label, and model the deformation of each vertex as dependent on the movement of corresponding part. We argue that this strategy tends to produce artifacts especially around joint positions where different parts connected with each other since vertex movement should have inter-dependencies with both local structures as well as full body articulation. 

Our work combine the advantages of both parametric models and learning-based ones in a way by representing a human body as distribution of points on its surface. Such a formulation allows us to bypass the ambiguities in mesh topology modeling since the basic structure of human body remains consistent. Meanwhile, our factorized latent representation facilitates structure-level control, outperforming previous works that only generate a full shape with single latent code and addressing the limitations of part-based representations.

\subsection{Denoising Diffusion Probabilistic Model}
Denoising Diffusion Probabilistic Models (DDPMs) are a kind of generative models that learn the distribution of data samples in a given dataset through a sequence of forward and reverse processes \cite{ho2020denoising,rombach2022high}. The forward process is defined as progressively injecting noise to the input with a variance schedule $\beta_1, \beta_2, \cdots, \beta_T$ until reaching an isotropic Gaussian distribution after sufficiently large $T$ steps:
\begin{equation}
    \begin{aligned}
        q(\mathbf{x}_t | \mathbf{x}_{t-1}) &=\mathcal{N}(\mathbf{x}_t ;
    \sqrt{1-\beta_t} \mathbf{x}_{t-1}, \beta_t \mathbf{I}), \\
    \quad q(\mathbf{x}_{1: T} | \mathbf{x}_0) &=\prod_{t=1}^T q(\mathbf{x}_t | \mathbf{x}_{t-1}).
    \end{aligned}
\end{equation}
The reverse process learns a parameterized transition kernel $\theta$ that inverts the forward diffusion process. In this way, we can synthesize novel data that follow certain distributions by initializing from random noise and sampling from the kernel from $t=T$ back to $t=0$ in an iterative fashion.
\begin{equation}
    \begin{aligned}
        p_\theta(\mathbf{x}_{t-1} | \mathbf{x}_t) &= \mathcal{N}(\mathbf{x}_{t-1} ; \mathbf{\mu}_\theta(\mathbf{x}_t, t), \mathbf{\Sigma}_\theta(\mathbf{x}_t, t)),\\
        p_\theta(\mathbf{x}_{0: T}) &= p(\mathbf{x}_T) \prod_{t=1}^T p_\theta(\mathbf{x}_{t-1} | \mathbf{x}_t),
    \end{aligned}
\end{equation}

Diffusion model's capability of modeling complex data distribution enables it to achieve remarkable performances in various research areas like 2D vision~\cite{dhariwal2021diffusion,rombach2022high,ho2022video,song2020score,preechakul2022diffusion,peebles2023scalable}, natural language processing~\cite{li2022diffusion,gong2022diffuseq,he2022diffusionbert}. Recently, the applications are extended to 3D vision tasks~\cite{hong2023lrm,poole2022dreamfusion,tang2023make,liu2023zero}, revealing its potential in generative modeling.

Most relevant to ours are the works applying latent diffusion to 3D shape generation task, including LION~\cite{vahdat2022lion}, SLIDE~\cite{vahdat2022lion}, DiffFacto~\cite{nakayama2023difffacto} and others~\cite{koo2023salad,luo2021diffusion}. Their success have brought us two key observations which highly inspire our work: (i) point clouds are naturally more suitable for diffusion models than other representations like meshes or voxels, because modeling the distribution of vertex positions is easier and more straightforward than modeling inherent interdependencies; (ii) to facilitate flexible manipulations, careful design of structure-aware latent representations, like segmented parts~\cite{nash2017shape,li2022editvae,hertz2022spaghetti,nakayama2023difffacto} and hierarchies~\cite{vahdat2022lion,lyu2023controllable}, are needed. However, most existing works cannot be effectively applied to 3D human modeling since they primarily focus on the modeling of static rigid objects like cars, airplanes, chairs, etc, while human body is non-rigid deformable, whose characteristics make previous latent representations inapplicable. This motivates us to take advantage of the traditional parameterized human representations and design a joint-aware latent diffusion pipeline suitable for modeling the distribution of shape variations. To our best knowledge, we are the first to explore related concepts for 3D human modeling and also achieve good results. 
\section{Methods}
\label{sec:method}

\begin{figure*}[t]
    \centering
    \begin{subfigure}[t]{0.75\linewidth}
        \centering
        \includegraphics[width=\textwidth]{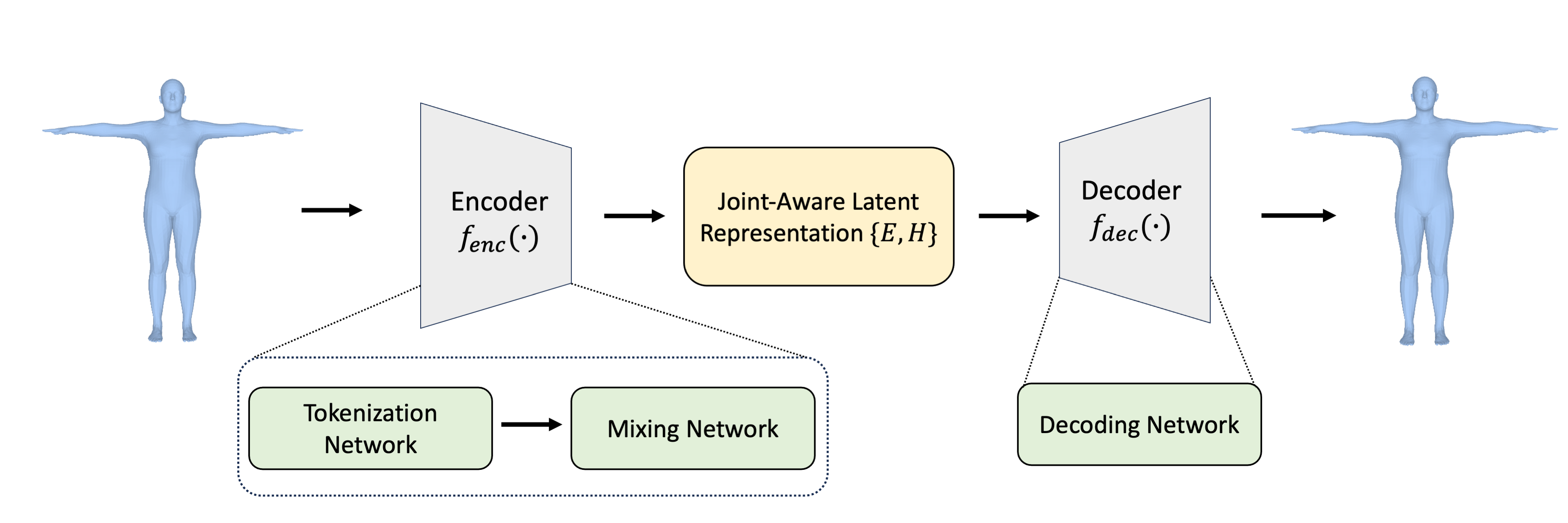}
        \caption{Overview of the autoencoder architecture.}
    \end{subfigure}
    \hfill
    \begin{subfigure}[t]{\linewidth}
        \centering
        \includegraphics[width=0.75\textwidth]{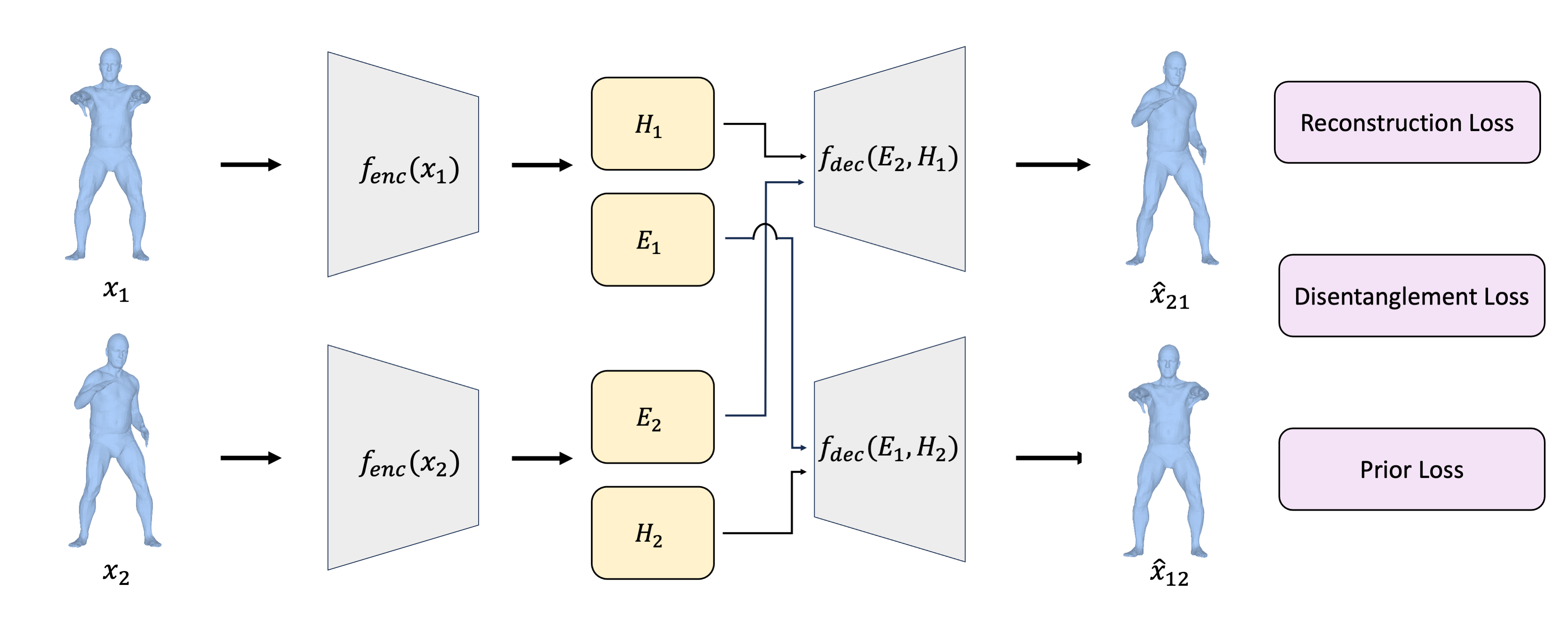}
        \caption{Joint-aware latent representation training pipeline.}
    \end{subfigure}
    \caption{A visual illustration of the autoencoder architecture that is used to train our joint-aware latent representation as well as its training pipeline. (a) The encoder $f_{enc}(\cdot)$, which consists of a tokenization network and a mixing network, maps a mesh human shape $x$ into the extrinsic parameters $E$ and intrinsic features $H$, and the decoder $f_{dec}(\cdot)$ aims to recover the original shape using the paired latents. (b) On top of typical reconstruction loss, we also employ a disentanglement loss to ensure the skeleton structure and the geometric details are independently preserved, as well as a prior loss to regularize a smooth latent space.}
    \label{fig:jade_pipeline}
\end{figure*}

\subsection{Overview}
\label{sec:overview}
Given the set of surface point clouds $\mathbf{X} \in \mathbb{R}^{N \times 3}$ of human meshes that consist of $N$ points with consistent connectivity, our goal is to learn a distribution $p(\mathbf{X})$ from which we can generate high-fidelity human bodies with fine-grained-level control. In the following sections, we first introduce the joint-aware latent representation design, which is core to JADE. Training is then performed in two stages - first, we implement a Transformer-based autoencoder to conduct representation learning through self-supervised setting; next, we build a cascaded diffusion pipeline on the factorized latent encodings to model their precise distributions and improve the generation quality.

\subsection{Joint-aware Latent Representation}
As raw point clouds are redundant to store and difficult to manipulate directly, we would like to encode them to a more compact latent representation that captures both low-level geometry and high-level semantic information. The primary objective that guide our design is that the latents should be structure-aware, i.e. each sub-structure of human body is encoded independently while their aggregation can still form a globally coherent shape.

To this end, we introduce a factorized representation of human shapes by partitioning them into $J$ different joints, as shown in Figure \ref{fig:joint-aware decomposition}, whose number agrees with the cardinality according to anatomical splits, and utilize joint-wise embeddings as our learning goal. The intuition behind this comes from the concept of skinning in parametric human modeling~\cite{loper2023smpl} which states that vertex positions on human bodies can be represented by a blend of multiple transformations associated with nearby joints. Therefore, we encode 3D human bodies into a set of joint-wise tokens $\mathbf{Z} = \{\mathbf{z}_i \in \mathbb{R}^{D_z}\}_{i=1}^J$, which can be further decomposed into two components. The first one is the skeleton structure $\mathbf{E} = \{\mathbf{e}_i \in \mathbb{R}^{3}\}_{i=1}^J$, referred to as extrinsics, characterized by joints positions and the second one are features attached to each joint $\mathbf{H} = \{\mathbf{h}_i \in \mathbb{R}^{D_h}\}_{i=1}^J$, identified as intrinsics, that model the interdependencies between joints and surface vertices to supplement detailed local surface geometry information. Our proposed shape factorization is then given as
\begin{equation}
\label{shape factorization}
    \begin{aligned}
        p(\mathbf{X}) = \prod_{i=1}^J p(\mathbf{z}_i) = \prod_{i=1}^J p(\mathbf{e}_i)p(\mathbf{h}_i | \mathbf{e}_i)
    \end{aligned}
\end{equation}

In contrast to common structure-aware latent designs that either decompose objects into predefined set of semantic parts~\cite{nakayama2023difffacto}, or use farthest point sampling (FPS) algorithm to sample sparse points as latent hierarchies~\cite{lyu2023controllable,vahdat2022lion}, our joint-aware representation has several advantages: (i) it addresses the critical problem that part segmentation labels are expensive to obtain, while also capable of adaptively decomposing human bodies into different semantically meaningful parts (ii) the extrinsic-intrinsic disentanglement makes our latent possess both geometric and semantic information, which facilitates flexible editing.

\subsection{Autoencoder Architecture}
In this section, we demonstrate how the aforementioned latent representation is learned in an autoencoding setup, whose architecture and training pipeline is illustrated in Figure \ref{fig:jade_pipeline}. The whole architecture comprises of three parts. The first module, the Tokenization Network, transforms a 3D human point cloud to a sparse set of $J$ tokens. The second component, the Mixing Network, augments these tokens with contextual information from each other, which reinforce their capabilities to remain aware of both global skeletal structure and local geometric details, thus generating the joint-aware latent representation. The final building block, the Decoding Network, reconstructs these latents back to original input point cloud without any external supervision. 

\noindent\textbf{Tokenization Network}
Our first objective is to obtain a consistent joint-level decomposition of a given 3D human shape, which forms the basis of the joint-aware representation. Given the surface points of a 3D human body, we map them to a global latent using PointNet~\cite{qi2017pointnet} and then split it into discrete set of tokens denoted as $\{\mathbf{f}_i \in \mathbb{R}^{D_z}\}_{i=1}^J$ with a simple MLP, where $D_z$ indicates the feature dimension. 

\noindent\textbf{Mixing Network}
Given the outputs from tokenization module, we then aim to extract the extrinsic and intrinsic information out of each token to achieve semantic-geometric disentanglement. We first project them onto high dimension vectors and then sum up with a learnable positional embedding, which indicates the relative position of local area each token should attend to. The resulting features $\{\mathbf{z}_i^0\}_{i=1}^J$ are fed to a sequence of standard transformer encoder blocks which applies the self-attention mechanism to integrate information across all embeddings and obtain the output features $\{\mathbf{z}_i^L\}_{i=1}^J$ where the superscript indicates which encoder block we're extracting the feature from. On top of that, each token is projected to two sets of parameters: extrinsic parameters represented by joint location $\mathbf{e}_i \in \mathbb{R}^{3}$, and intrinsic surface geometry information $\mathbf{h}_i \in \mathbb{R}^{D_h}$. 
\begin{equation}
\label{encoder mlp}
    \begin{aligned}
        \mathbf{e}_i &= \text{MLP}_{e}(\mathbf{z}_i^L) \\
        \mathbf{h}_i &= \text{MLP}_{h}(\mathbf{z}_i^L) \\
    \end{aligned}
\end{equation}
The combination of tokenization network and mixing network constitutes the encoder part of our framework $f_{enc}(\cdot): \mathbb{R}^{N \times 3} \rightarrow \mathbb{R}^{J \times (3 + D_h)}$.

\noindent\textbf{Decoding Network}
Our decoding network $f_{dec}(\cdot): \mathbb{R}^{J \times (3 + D_h)} \rightarrow \mathbb{R}^{N \times 3}$ also utilizes a Transformer-based architecture, takes the latent representations as input and reconstructs them back to the original shape. We separate extrinsic parameters $\mathbf{E} \in \mathbb{R}^{J \times 3}$ from intrinsic features $\mathbf{H} \in \mathbb{R}^{J \times D_h}$ in the latents and treat skeleton structures as an external condition for better guidance during the decoding process. The conditioning strategy is implemented by simply concatenating the two components before feeding in the network. After the input concatenation and summing up with the shared learnable positional embedding used in the mixing network, we can both anchor the absolute spatial location that corresponding token should attend to and preserve their relative order in the input sequence. The output features obtained after passing through transformer blocks are connected with a MLP head and convert them to a complete point cloud object, which concludes the autoencoding pipeline.

\subsubsection{Training Losses}
To successfully complete the training pipeline and obtain the joint-aware latent representations, we introduce the loss function as follows:
\begin{equation}
    \begin{aligned}
        \mathcal{L}_{total} &= \mathcal{L}_{rec} + \mathcal{L}_{dis} + \mathcal{L}_{prior}
    \end{aligned}
\end{equation}

\noindent\textbf{Reconstruction loss}. In order to reconstruct the original human mesh as accurately as possible, we adopt the geometric reconstruction loss in both vertex-level and joint-level
\begin{equation}
\label{eq:rec}
    \begin{aligned}
        \mathcal{L}_{rec} &= \mathcal{L}_{verts} + \lambda_{j}\mathcal{L}_{joints} \\
        &= \|\mathbf{X} - \hat{\mathbf{X}}\| + \lambda_{j} \|\mathbf{J} - \hat{\mathbf{J}}\|
    \end{aligned}
\end{equation}
where $\mathbf{X}$, $\mathbf{J}$ are the surface vertices and joints locations of the original mesh and $\hat{\mathbf{X}}$, $\hat{\mathbf{J}}$ are the ones from the reconstructed mesh. Recall that extrinsic parameters $\mathbf{E}$ are simply the joint positions so that $\hat{\mathbf{J}} = \mathbf{E}$. These losses force the reconstructed mesh to be close to the original one so that the latents can capture as much information as possible. In addition, the joint supervision loss also enables tokens to aggregate geometric information around specific joints, which to some extent facilitates the joint-aware design.

\noindent\textbf{Disentanglement loss}. After applying the reconstruction loss, our framework is already capable of accurate reconstruction, but its latent space is still entangled as geometric information might flow into semantic features. To decouple extrinsic and intrinsic components from the latent representation, we apply a disentanglement loss, as introduced in~\cite{zhou2020unsupervised}, during training:
\begin{equation}
\label{eq:dis}
    \begin{aligned}
        \mathcal{L}_{dis} &= \lambda_{c} \mathcal{L}_{cross} \\
        \mathcal{L}_{cross} &= \|\hat{\mathbf{X}}_{12} - \mathbf{X}_1\| + \|\hat{\mathbf{X}}_{21} - \mathbf{X}_2\| \\
        \hat{\mathbf{X}}_{ij} &= f_{dec}(\mathbf{E}_i, \mathbf{H}_j) \\
    \end{aligned}
\end{equation}
More specifically, we sample a mesh tuple $\{\mathbf{X}_1, \mathbf{X}_2\}$ from the same subject but with different poses during every training iteration. The encoder takes mesh $\mathbf{X}_i, i=1,2$ as input and outputs corresponding extrinsic and intrinsic vectors $(\mathbf{E}_i, \mathbf{H}_i), i=1,2$. The cross consistency loss is computed by swapping the intrinsics features of two deformations of the same subject to reconstruct each other.

\noindent\textbf{Prior loss}. We further add a weighted Kullback–Leibler divergence loss between the distribution of intrinsic features $p(\mathbf{H})$ and the standard normal distribution $\mathcal{N}(\mathbf{0}, \mathbf{I})$:
\begin{equation}
    \begin{aligned}
        \mathcal{L}_{prior} &= \lambda_{kl} \left[ D_{KL}(p(\mathbf{H}) || \mathcal{N}(\mathbf{0}, \mathbf{I})) \right]
    \end{aligned}
\end{equation}
This regularization term aims to encourage the latent space to be simple and smooth, so that we can perform generation and interpolation through sampling data points in this space.

\subsection{Latent Cascaded Diffusion}
\begin{figure*}[t]
    \begin{center}
    \includegraphics[width=0.9\linewidth]{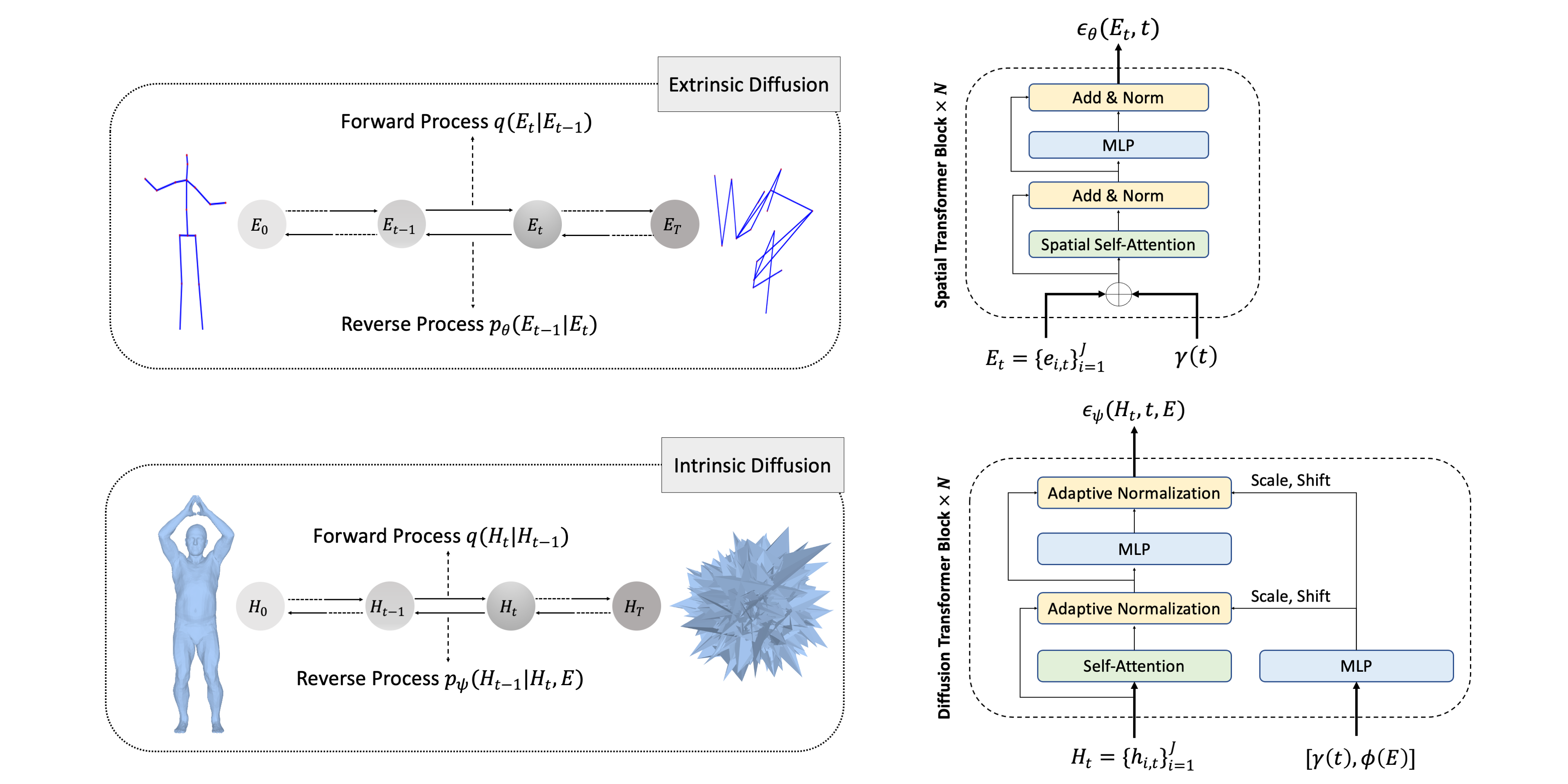}
    \caption{A visual illustration of the diffusion pipeline, where two cascaded diffusions are presented, one for extrinsic parameters $\mathbf{E} = \{\mathbf{e}_i\}_{i=1}^J$ and the other for intrinsic features $\mathbf{H} = \{\mathbf{h}_i\}_{i=1}^J$. In the first phase, we use a time-conditioned spatial transformer to handle diffusion on set data, and in the second phase, we utilize a DiT to handle more complex conditioning, where the concatenation of encoded timestamp $\gamma(t)$ and extrinsics outputs from phase one $\phi(\mathbf{E})$ is fed to the adaptive normalization layers to modulate the diffusion process.}
    \label{fig:diffusion}
    \end{center}
\vspace{-1em}
\end{figure*}

In principle, after finish training the autoencoder, we could use the standard Gaussian’s priors to sample latent representations and generate new human body shapes. However, such simple priors may not accurately model the distribution of latent space and thus produce low-quality generated samples, whose phenomenon is also known as prior hole problem~\cite{vahdat2021score,takahashi2019variational,vahdat2020nvae}. This motivates us to resort to training a highly expressive DDPM on the latent space to model its precise distribution. 

To properly handle the diffusion of latent representations, we utilize the characteristics of extrinsic-intrinsic disentanglement and apply a cascaded diffusion scheme according the shape factorization in Equation \ref{shape factorization}. The architectures of two diffusion models are illustrated in Figure \ref{fig:diffusion}. The first DDPM learns the distribution of extrinsics $p(\mathbf{E})$ with a Transformer-based noise prediction network $\boldsymbol{\epsilon}_{\theta}$~\cite{ji2024unsupervised}, providing coarse-level skeletal structure information of the human body. The second one uses a DiT-based network $\boldsymbol{\epsilon}_{\psi}$~\cite{peebles2023scalable} to learn the conditional distribution of intrinsics $p(\mathbf{H} | \mathbf{E})$ conditioned on the skeleton structure $\mathbf{E}$ to capture finer geometric details. Specifically, the encoded timestamp $\gamma(t)$ and extrinsic embedding $\phi(\mathbf{E})$ are concatenated and fed into the adaptive normalization layers such as AdaLN~\cite{perez2018film} so that it can learn a scale and translation factor from the condition signals to adaptively modulate the network output. Both the noise prediction networks $\boldsymbol{\epsilon}_{\theta}$ and $\boldsymbol{\epsilon}_{\phi}$ are trained with the same variational bound loss as follows:
\begin{equation}
    \begin{aligned}
        \mathcal{L}_{\mathbf{E}}(\theta) &= \mathbb{E}_{t,\mathbf{E}, \boldsymbol{\epsilon}} \left[ \left\| \boldsymbol{\epsilon} - \boldsymbol{\epsilon}_\theta \left( \mathbf{E}_{t}, t \right) \right\|^2 \right] \\
        \mathcal{L}_{\mathbf{H}}(\psi) &= \mathbb{E}_{t,\mathbf{H}, \boldsymbol{\phi}} \left[ \left\| \boldsymbol{\epsilon} - \boldsymbol{\epsilon}_\phi \left( \mathbf{H}_{t}, t, \mathbf{E} \right) \right\|^2 \right]
    \end{aligned}
\end{equation}
where the subscription $t$ indicates the attributes results after $t$-step forward process of adding Gaussian noise.
\section{Experiments}
\label{sec:exp}

In this section we evaluate the performance of JADE in various human-centric applications, including human shape representation, editing and generation. We first elaborate the dataset configurations, evaluation metrics and implementation details of our approach, and then demonstrate the advantages of JADE for each application compared with prior works. Additionally, we conduct thorough ablation studies to delve into the key elements contributing to our model’s performance.

\subsection{Experiment Setups}
\subsubsection{Datasets}
\noindent\textbf{DFAUST}~\cite{bogo2017dynamic} captures 14 different body motion sequences (e.g., hips, running, and jumping) for each of the 10 human subjects and register a reference template mesh with same topology to these raw scans. We split off two dynamic performances, i.e.,one-leg jump and chicken wings to conduct testing, which results in a training set with 29,005 samples and a testing set with 3,919 samples.

\noindent\textbf{SPRING}~\cite{yang2014semantic} provides a comprehensive collection of 3D meshes with a rough A-pose registered from the CAESAR dataset~\cite{robinette1999caesar} using a non-rigid deformation algorithm. We preprocess the SPRING dataset in advance to make its data format compatible with the more commonly-used SMPL model so that the mesh connectivity can be consistent. For the subsequent experiments, the SPRING dataset is randomly split into 2743 training and 305 test meshes.

\noindent\textbf{AMASS}~\cite{mahmood2019amass} aggregates motion capture data from various sources, transforming it into a unified format with 3D human body meshes using the SMPL model. It provides detailed 3D motion data, including SMPL parameters, joint angles, and shape coefficients. In our experiments, we adhere to the same train-test partition as previous works~\cite{pavlakos2019expressive,lu2023dposer,tiwari2022pose} but evenly extract one-tenth of the original dataset for the convenience of training, resulting in roughly two million data samples.

\subsubsection{Evaluation Metrics}
To comprehensively evaluate our framework across various tasks, we adopt task-specific metrics accordingly. For representation ability evaluation, we simply utilize Mean Per Vertex Position Error (MPVPE), which is calculated as the average Euclidean distance between corresponding vertices in the ground truth and its reconstruction as metrics. For generation quality evaluation, both the diversity and fidelity of generated human bodies needs to be considered. To assess generation diversity, we utilize Average Pairwise
Distance (APD)~\cite{aliakbarian2020stochastic}, defined as mean joint distance between all pairs of samples, as metrics, and for realism evaluation, we employ Self-Intersection rates (SI)~\cite{lu2023dposer}, which is defined as the average percentage of self-intersecting faces in a batch of 3D meshes.

\subsubsection{Implementation Details}
As mentioned in Section \ref{sec:overview}, training of JADE is performed in two stages. For the latent representation training, we set the number of joints $J = 24$, the feature dimension $D_z = D_h = 128$, and the autoencoder is trained using AdamW~\cite{loshchilov2017decoupled} optimizer with batch size 256, learning rate $10^{-3}$. For the diffusion process, we follow conventional DDPM training strategy, where the maximum iteration is set as $T=1000$, the timestamp $t$ is uniformly sampled from $[1, T]$, and the variances of added noises are configured to linearly increase from $\beta_1=10^{-4}$ to $\beta_T=0.02$.  We also record the exponential moving average of the DDPM parameters along the training trajectory and the ratio is set to 0.9999. All 3D human shapes are normalized to pelvis-related coordinates. Our experiments are performed on one NVIDIA 3090 GPU, and the training takes about 2 days to complete. For all baselines, we train them using their released codebase and follow the default setting.

\subsection{Human Shape Reconstruction}
In this section, we validate the representation ability of our approach on DFAUST and SPRING datasets. We compare the results against various kinds of methods, including classical parametric models~\cite{loper2023smpl,bogo2016keep,pavlakos2019expressive}, spectral-based approach~\cite{ranjan2018generating}, spiral-based method~\cite{bouritsas2019neural}, and disentangled representations~\cite{zhou2020unsupervised,sun2023learning}. As shown in Table \ref{tab:reconstriction}, our approach outperforms the methods in all categories with high reconstruction precision, demonstrating the effectiveness of the joint-aware latent representation. Unlike mesh-based representations that use thorough topological information during training, our approach is point cloud-based, which only utilize vertex positions to model the surface geometry. Figure \ref{fig:error_map} also visualizes some qualitative results and their error maps so that we can observe that JADE exhibits better reconstruction accuracy especially for complex geometric details.
\begin{table}[t]
\setlength{\abovecaptionskip}{0.2cm}
\centering
    \begin{tabular}{ l  l  c  c  c  c }
    \toprule
        \multirow{2}{*}{ Mode } & \multirow{2}{*}{ Method } & \multicolumn{2}{c}{ MPVPE $\downarrow$} \\
        & & DFAUST & SPRING \\
    \midrule
        \multirow{4}{*}{ Par. } 
        & SMPL~\cite{loper2023smpl} & 24.67 & - \\
        & SMPL-H~\cite{romero2022embodied} & 21.38 & -  \\
        & SMPL-X~\cite{pavlakos2019expressive} & 18.71 & - \\
    \midrule
        \multirow{5}{*}{ Lrn. } 
        & COMA~\cite{ranjan2018generating} & 30.83  & 45.53 \\
        & Neural3DMM~\cite{bouritsas2019neural} & 16.46 & 15.54\\
        & UnsupShapePose~\cite{zhou2020unsupervised} & 9.61 & \underline{13.65}\\
        & SemanticHuman~\cite{sun2023learning} & \underline{5.70} & 18.76  \\
    \midrule
        \rowcolor[gray]{0.85}
        & \textbf{JADE} & \textbf{5.47} & \textbf{12.85} \\
    \bottomrule
    \end{tabular}

\caption{Quantitative reconstruction results on DFAUST and SPRING datasets. \textit{Par.} stands for parametric model and \textit{Lrn.} means learning-based methods. \textit{MPVPE} is the mean position error per vertex. $-$ : not supported for this dataset. The best and second-best results are highlighted in bold and underline formats.}
\label{tab:reconstriction}
\end{table}

\begin{figure}[t]
    \begin{center}
    \setlength{\abovecaptionskip}{-0.1cm}
    \includegraphics[width=\linewidth]{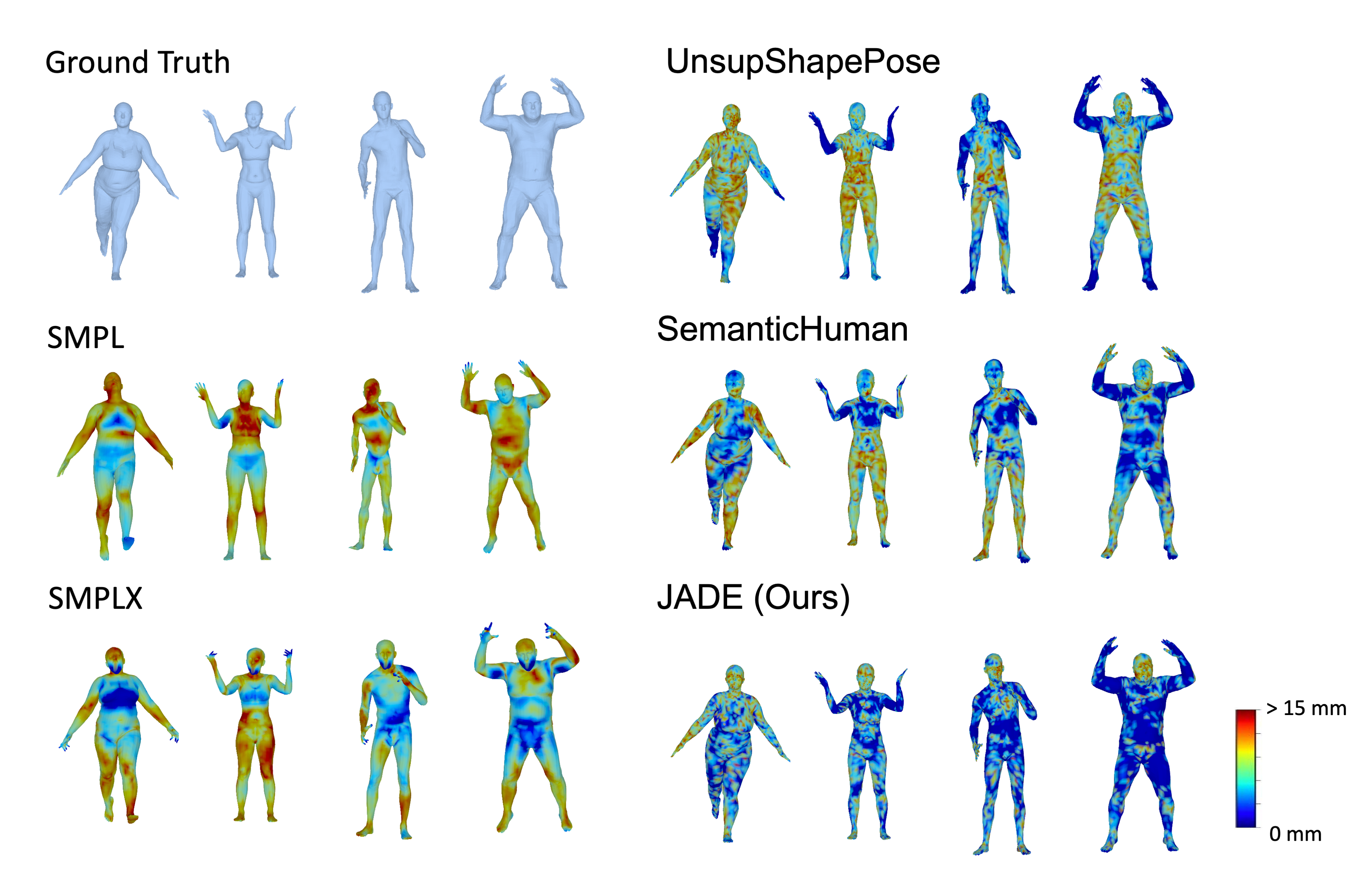}
    \caption{Qualitative visualization results on DFAUST dataset, showing the color coding of the MPVPE error of the reconstructions produced by our JADE framework and baseline methods\cite{pavlakos2019expressive,loper2023smpl,zhou2020unsupervised,sun2023learning}. The error maps show that our method has better reconstruction accuracy.}
    \label{fig:error_map}
    \end{center}
\vspace{-2em}
\end{figure}

\subsection{Human Shape Editing}
\begin{figure}[t]
    \begin{center}
    \setlength{\abovecaptionskip}{-0.1cm}
    \includegraphics[width=\linewidth]{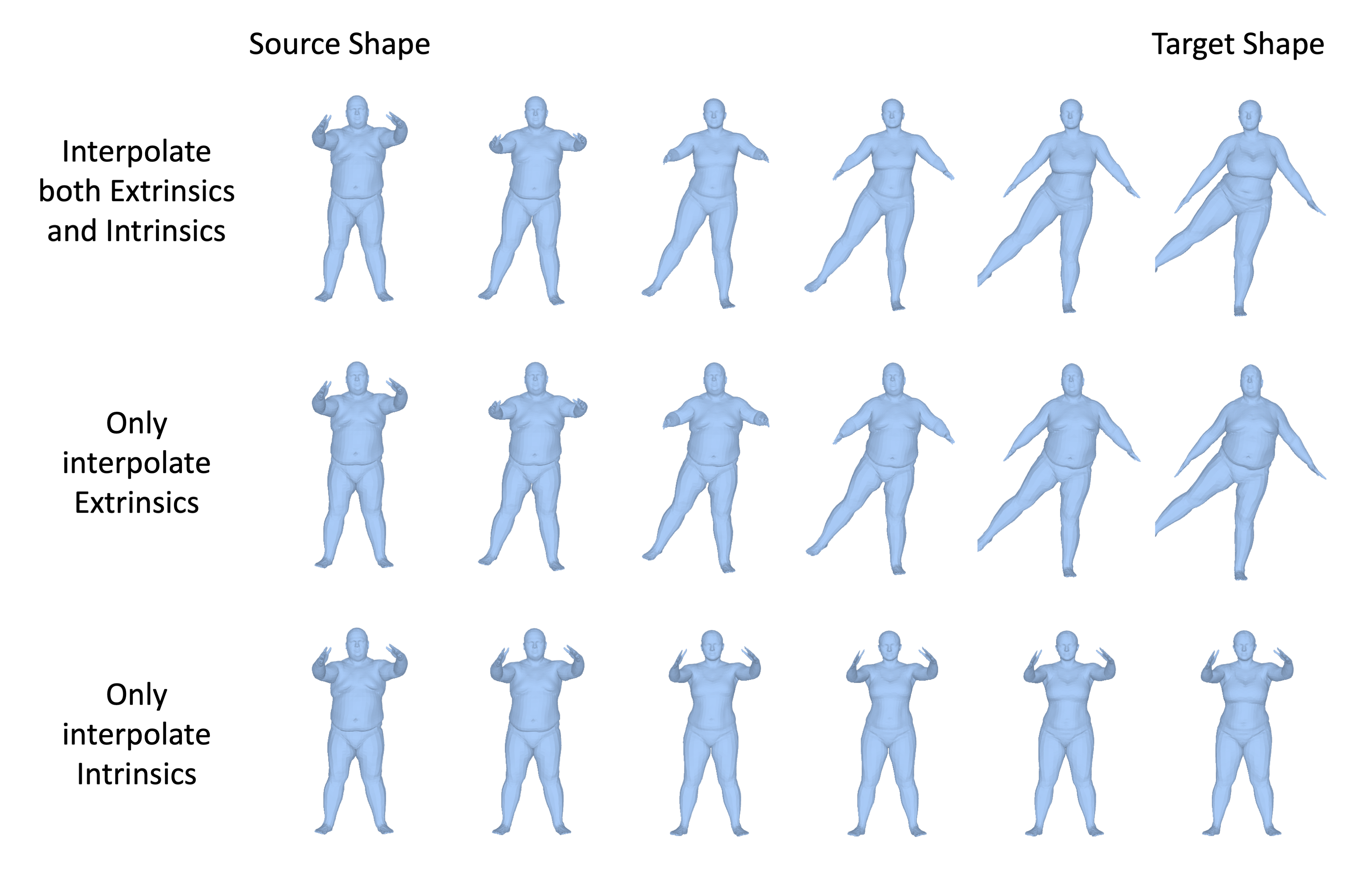}
    \caption{Qualitative example of transferring a given human character to a new identity with different body posture through interpolation, indicating that JADE can edit both the extrinsics and intrinsics components in the latent space independently to achieve desired human body movements.}
    \label{fig:interpolation}
    \end{center}
\end{figure}

In this section, we demonstrate that the joint-aware latent representation has great flexibility to perform human shape editing through transfer experiments. The top row of Figure \ref{fig:interpolation} shows an example of shape transfer, who aims to transform from a given human character to a new identity with different body postures. JADE can complete this task through interpolation. Thanks to the disentangled latent design, we can interpolate both the extrinsic parameters and intrinsic features correspond to each joint to achieve desired human body movements. We also show the influence of extrinsic-intrinsic pair in the latent space decomposition by only interpolating each component respectively. From the second and third row of Figure \ref{fig:interpolation}, we can see that extrinsics control the overall skeleton structure and posture of the human body while the intrinsics control the local geometry.

\begin{table}[t]
\setlength{\abovecaptionskip}{0.2cm}
\centering
    \begin{tabular}{ l  c  c }
    \toprule
     Sample source  & APD $\uparrow$  & SI $\downarrow$ \\
    \midrule
        AMASS~\cite{mahmood2019amass}  & 15.44 & 0.79 \\
    \midrule
        GMM~\cite{bogo2016keep} & \underline{16.28} & 1.54 \\
        VPoser~\cite{pavlakos2019expressive} & 10.75 & 1.51 \\
        PoseNDF~\cite{tiwari2022pose} & \textbf{18.75} & 1.97 \\
        DPoser~\cite{lu2023dposer} & 14.28 & \underline{1.21} \\
    \midrule
        \rowcolor[gray]{0.85}
        \textbf{JADE} & 14.95 & \textbf{1.05} \\
    \bottomrule
    \end{tabular}
\caption{Quantitative generation metrics on AMASS dataset.}
\label{tab:pose generation}
\end{table}

\begin{table*}[t]
\setlength{\abovecaptionskip}{0.2cm}
\centering
    \begin{tabular}{ c | c | c | c c | c c }
    \toprule
         & \multirow{2}{*}{ Intrinsics Dimension } & \multirow{2}{*}{ Condition Mechanism } & \multicolumn{2}{c|}{ Loss function } & \multicolumn{2}{c}{ MPVPE $\downarrow$ } \\
         & & & $\mathcal{L}_{joint}$ & $\mathcal{L}_{dis}$ & DFAUST & SPRING \\
    \midrule
        Full Architecture & 128 & concat & \checkmark & \checkmark & 5.47 & 12.85 \\ 
    \midrule
        \multirow{5}{*}{ Variant 1 }
        & 16 & concat & \checkmark & \checkmark & 7.85 & 19.01 \\
        & 32 & concat & \checkmark & \checkmark & 6.37 & 15.12 \\
        & 64 & concat & \checkmark & \checkmark & 5.83 & 13.55 \\
        & 256 & concat & \checkmark & \checkmark & 6.29 & 14.80 \\
        & 512 & concat & \checkmark & \checkmark & 8.11 & 19.04 \\
    \midrule
        \multirow{2}{*}{ Variant 2 } 
        & 128 & add & \checkmark & \checkmark & 5.61 & 13.43 \\
        & 128 & cross-attention & \checkmark & \checkmark & 9.89 & 24.33 \\
    \midrule
        \multirow{2}{*}{ Variant 3 } 
        & 128 & concat & & \checkmark & 6.01 & 14.12 \\
        & 128 & concat & \checkmark & & 6.64 & - \\
    \bottomrule
    \end{tabular}

\caption{Ablation studies on the intrinsics dimension, condition mechanism and the impact of different loss functions. We show the experiment results under reconstruction setting and report the performance of MPVPE on DFAUST and SPRING dataset.}
\label{tab:ablation rec}
\end{table*}

\subsection{Human Shape Generation}
We use AMASS dataset to train JADE solely for unconditional human shape generation and compare it against various human body generation methods such as classical GMM~\cite{bogo2016keep}, VPoser~\cite{pavlakos2019expressive}, Pose-NDF~\cite{tiwari2022pose}, DPoser~\cite{lu2023dposer}. The quantitative and qualitative results are summarized in Table \ref{tab:pose generation}. Following the strategy of previous works, we randomly sample 500 different human shapes and evaluate their diversity and realism through APD and SI metrics. Compared with ground truth from AMASS dataset, Pose-NDF and classicl GMM exhibit both high APD and SI, suggesting more divergence from training samples but resulting in more unrealistic human shapes. On the contrary, VPoser and DPoser shows lower scores in both metrics, indicating they prefer realism over diversity. JADE proves to be a good trade-off between these two criteria, both numerically and visually.

\subsection{Ablation Studies}
To validate our architecture design and training settings, we compare the final model of the latent representation with several types of variants and evaluate their performances under the reconstruction experiment setting. The results of the ablation studies are summarized in Table \ref{tab:ablation rec} and we'll further discuss the details below.

\noindent\textbf{Effect of intrinsics feature dimension.} The first type of variants investigate how the dimension of intrinsic features affect the representation ability. We can see that as the feature dimension increases, the reconstruction error decreases at first, reaching a minimum at 128, and then increases. This reveals that enlarging the latent dimension in some extent contributes to the representation ability, but will introduce redundancy, similar to overfitting phenomena, and hamper its generalization expressivity once exceeds a threshold.

\noindent\textbf{Effect of conditioning mechanism.} The second type of variants examine multiple conditioning methods, including concatenation, addition, and cross-attention, to inject skeleton structure information in the decoding process. For all experiments, we first project the skeleton structure onto a high-dimensional embedding, after which these embeddings are fused into the decoding network. The best conditioning mechanism is concatenating the skeleton embedding and intrinsics latent feature, which provides a simple and effective way to guide the latent decoding.

\noindent\textbf{Effect of loss functions.} The third type of variants explore the influence of loss functions, where the network architecture remains the same, but we omit the joint supervision loss $\mathcal{L}_{joint}$ in Equation \ref{eq:rec} and disentangled loss $\mathcal{L}_{cross}$ in Equation \ref{eq:dis} from the training objective. Although these reduced variants achieve slightly worse reconstruction performances, the characteristics of latent disentanglement, structure-level control and the quality of editable human shapes degrades significantly, which assure the importance of our losses design.

\section{Conclusion}
In this work, we introduce JADE, a generative framework trained on surface point clouds for 3D human modeling. Our key insight is a joint-aware latent representation that decomposes human bodies into skeleton structures and local surface geometries. The disentangled design enables geometric and semantic interpretation, facilitating users with flexible controllability. We also present a cascaded pipeline to generate coherent and plausible human shapes under our proposed decomposition. Extensive experiments conducted on public datasets demonstrates the effectiveness of JADE framework in multiple downstream tasks. Currently, JADE is limited on fixed topology human shape modeling and can not directly generate textured shapes. To bypass the underlying mesh topology restriction, a promising extension would be to incorporate neural implicit representation to generate more smooth shapes. In addition, combining with image-based training and differentiable rendering techniques to also synthesize textures would also be a potential future research direction.
{
    \small
    \bibliographystyle{ieeenat_fullname}

}

\end{document}